\documentclass{article}

\usepackage{microtype}
\usepackage{graphicx}
\usepackage{subfigure}
\usepackage{booktabs} 

\usepackage{hyperref}

\usepackage[accepted]{icml2025}
\hypersetup{pdfsubject={}}
\usepackage{amsmath}
\usepackage{amssymb}
\usepackage{mathtools}
\usepackage{amsthm}

\usepackage[capitalize,noabbrev]{cleveref}

\theoremstyle{plain}

\theoremstyle{definition}

\theoremstyle{remark}

\usepackage[textsize=tiny]{todonotes}

\begin{document}

\makeatletter
\renewcommand{\Notice@String}{}
\makeatother

\twocolumn[
\icmltitle{Layer-Level Control of Language Model Execution in Agent Frameworks}





\begin{icmlauthorlist}
\icmlauthor{Ran Guan}{}
\icmlauthor{Yiqi Lin}{}
\icmlauthor{Xiangxi Tian}{}
\icmlauthor{Yige Luo}{}
\icmlauthor{Mi Zhou}{}
\end{icmlauthorlist}

\begin{center}
    2012 Laboratories, Huawei Technologies
\end{center}


\icmlcorrespondingauthor{Ran Guan}{guanran@huawei.com}

\icmlkeywords{Agent Framework}

\vskip 0.3in
]




\begin{abstract}
Agent frameworks typically orchestrate language models through prompt-and-response interfaces. We propose a harness design that exposes intermediate model states and layer-level execution controls to the agent runtime. By connecting state collection, policy fitting, and execution, the design enables policies to allocate computation within and across models based on intermediate states. We demonstrate the approach with a prototype that learns stitching bridges and state-conditioned execution policies from paired model states, then applies them during inference to select among \texttt{continue}, \texttt{exit}, and \texttt{stitch}. For dispatcher supervision, we select the least-computation action whose full-vocabulary next-token prediction matches the source model's full-depth prediction, using \texttt{continue} when neither early action matches. We train these components on a frozen proxy corpus of 5,000 typed-decision items authored using source texts sampled from WildChat conversations, then evaluate the prototype on JevBench, measuring task accuracy, agreement with full-model predictions, and end-to-end latency including runtime overhead. Preliminary results for two model pairs show mean-latency reductions of approximately 15\% and 11\% while maintaining task accuracy comparable to full-model inference.

\end{abstract}

\section{Introduction}
\label{introduction}

Agent frameworks have become increasingly capable at deciding when to call a language model, which model to use, and what to do with its response. The model call itself usually remains a black box: the harness submits a prompt and receives an answer, while computation proceeds inside the model under the inference backend’s control. A language model need not be treated as an indivisible unit of computation. Pay-as-you-go inference offers an alternative: the framework uses intermediate model states to decide how much computation to allocate as inference proceeds.

\begin{figure}[t]
    \centering
    \includegraphics[width=\columnwidth]{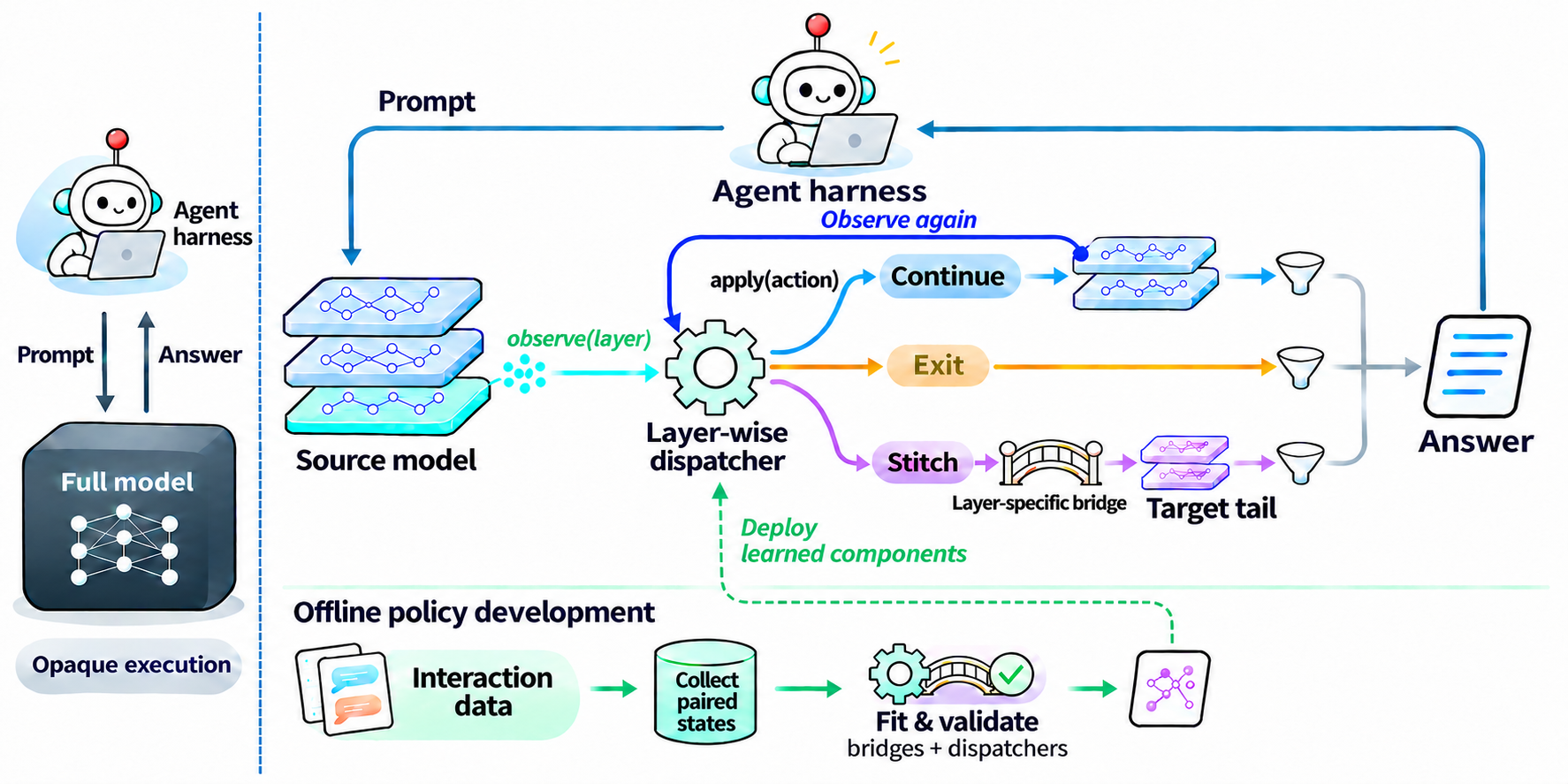}
    \caption{Conventional agent frameworks interact with a language model through opaque prompt--response calls (left). Our design exposes intermediate states and execution controls at layer boundaries, allowing the runtime to continue inference, exit early, or stitch into a smaller model; collected states support offline policy fitting and validation (right).}
    \label{fig:salami-design}
\end{figure}

Prior work has explored mechanisms such as early exit \citep{elhoushi2024layerskip}, model stitching \citep{hu2025stitchllm}, and speculative decoding \citep{li2026eagle,zhao2026tracespec} to reduce inference costs. These works motivate a complementary design question: how can an agent harness collect intermediate states, fit policies from interaction traces, and apply different execution controls through a common interface? We envision an agent framework that exposes layer-level states and execution actions to the harness runtime. With such a framework, prompts, responses, and their associated intermediate states can be collected during use to learn execution policies suited to the observed workload. At runtime, these policies can enable pay-as-you-go inference, allowing the framework to allocate computation at layer boundaries according to task requirements and observed states.

This short report demonstrates a prototype of such a framework. It provides a dedicated decision-model call for tasks that require selecting an answer from a fixed set of legal labels. As users interact with the agent, these calls and their intermediate states can accumulate into a corpus for developing and validating computation-allocation policies and their supporting components. To study this process under controlled conditions, we use 5,000 decision items constructed from WildChat, a dataset of real-world user–model conversations \citep{zhao2024wildchat}, in place of an ongoing interaction stream. Preliminary results suggest that tighter integration of the agent framework, inference engine, and model provides a promising foundation for exploring richer execution policies and computation–quality trade-offs.

\section{Framework Design}
\label{framework}

Conventionally, an agent framework invokes a language model and receives the resulting response through an interface such as \texttt{response = LLM(prompt)}. For simplicity, we model the language model as a stack of transformer layers. Our framework extends this interface with two abstract operations at configured layer boundaries:
\begin{itemize}
    \item \texttt{state = observe(layer)} exposes the intermediate state after the specified layer for inspection and optional storage.
    \item \texttt{apply(action)} controls how inference proceeds from the observed state.
\end{itemize}

For this report, we consider three execution actions. At each configured layer boundary, a runtime policy selects an action based on the observed state: \texttt{continue} executes the next layer of the original model and returns control to the runtime at the next configured boundary; if inference reaches the final layer, the model's output head produces the prediction; \texttt{exit} applies the original model's final normalisation and output head to the current state, then returns the highest-scoring legal label; and \texttt{stitch} maps the current state into the representation expected by another model, then resumes execution through that model's remaining layers and output head. The action set is extensible; we focus on these three actions in this prototype. Policy evaluation runs within the framework runtime and does not require additional language-model calls.

Let $c_s$ and $c_t$ denote the average execution costs per transformer block in the source and target models, respectively. Exiting at source layer $\ell$ in an $N$-layer model saves approximately $(N-\ell)c_s$. Stitching from source layer $\ell$ to target layer $j$ in an $M$-layer model costs approximately $\ell c_s + (M-j)c_t$, in addition to the stitching-map and runtime overheads. Because target-model blocks may be cheaper, stitching can reduce latency even when it executes many remaining blocks.

Both actions risk disagreement with full-model inference, so an effective execution policy must trade latency against prediction agreement and task quality. Our framework supports this trade-off at runtime while collecting intermediate states and execution outcomes. These data can be analysed offline to fit and evaluate policies across different operating points, helping identify a suitable operating point between computation, latency, and task quality for a target workload.

This report studies these questions in a controlled typed-decision setting. We assume that the framework provides a dedicated model call for decision tasks with a fixed set of legal answers. Given a prompt $x$ and candidate labels $\mathcal{Y}$, the call returns the label with the highest score under the selected execution path, $\hat{y} = \arg\max_{y \in \mathcal{Y}} s(y \mid x)$, where $s(y \mid x)$ denotes that path's score for candidate label $y$. This interface provides a common prediction target for comparing intermediate exits, stitched executions, and full-model inference.

\section{Experimental Setup}
\label{experiment}

We address two implementation questions in this section: how to learn a map that transfers an intermediate representation from a source model into a smaller target model, a stitching direction explored by prior work \citep{hu2025stitchllm}; and how to use the observed state to decide whether to continue computation, exit early, or apply the stitching map. \textbf{Our goal is not to improve task accuracy beyond the source model, but to preserve comparable benchmark accuracy while reducing inference computation and latency.} We train the bridges and dispatchers using agreement with the source model's full-depth predictions, preferring the least-computation action that matches them. We use a proxy dataset designed to approximate decision prompts and intermediate states accumulated through agent use to train the stitching maps and action policies, then evaluate the learned policies on a separate benchmark for task accuracy, model agreement, and latency.

Specifically, we construct a proxy corpus of 5,000 typed decision items from source texts selected from a pinned snapshot of WildChat, a dataset of real-world user--assistant conversations \citep{zhao2024wildchat}. Appendix~\ref{app:wildchat-selection} describes the source-selection procedure. We use the source texts to author 2,000 yes/no, 2,000 multiple-choice, and 1,000 scoring items; these are derived from WildChat conversations rather than native WildChat decision prompts. Appendix~\ref{app:decision-authoring} details the authoring prompts and validation procedure. We stratify the items by task type into 4,500 training and 500 validation examples. Notice that gold answers are not generated for this proxy corpus; it serves as a proxy for the decision-prompt workload used in policy development, rather than as a benchmark of task-solving accuracy.

For this study, we use Qwen3-8B, which has 36 transformer layers and a hidden dimension of 4,096, as the source model, and Qwen3-0.6B and Qwen3-1.7B as the target models for stitching. Both targets have 28 layers. Qwen3-0.6B has a hidden dimension of 1,024, and Qwen3-1.7B has a hidden dimension of 2,048. We run each of the 5,000 decision prompts through the source model and both target models and record the intermediate state after every layer. For a prompt of length $T$, the state after layer $\ell$ is $H_{\ell}^{(8\mathrm{B})} \in \mathbb{R}^{T \times 4096}$ for Qwen3-8B, $H_{\ell}^{(0.6\mathrm{B})} \in \mathbb{R}^{T \times 1024}$ for Qwen3-0.6B, and $H_{\ell}^{(1.7\mathrm{B})} \in \mathbb{R}^{T \times 2048}$ for Qwen3-1.7B; each row is the hidden representation of one prompt token.

As an initial heuristic, we compare the prediction at each layer with the source model's full-depth prediction on the proxy corpus. Matching exits are sparse before layer 23 and become more frequent overall at deeper layers (Appendix~\ref{app:exit-layer-agreement}). Although exiting earlier could save more source-model computation, these results suggest that exits before layer 23 are unlikely to preserve strong agreement with the full-depth prediction. We therefore use layer 23 as the earliest candidate boundary when evaluating early-exit policies. 

For \texttt{stitch} actions, we train a separate bridge $B_{L}$ for each configured source boundary $L$ and each target. It maps the Qwen3-8B state after layer $L$, $H_{L}^{(8\mathrm{B})}\in\mathbb{R}^{T\times 4096}$, to the target representation after layer $L-8$. The image of the bridge lies in $\mathbb{R}^{T\times 1024}$ for Qwen3-0.6B and in $\mathbb{R}^{T\times 2048}$ for Qwen3-1.7B. Because the source and both targets use the same tokenisation, the mapping is position-preserving: for each prompt-token position $u$, the bridge maps the source vector $H_{L}^{(8\mathrm{B})}[u,:]$ to the target vector at the same position. The mapped sequence is then processed by the remaining target-model layers using the original positions and causal attention mask; for the typed-decision task, the final prompt position is used to score the legal labels. We use an end-alignment heuristic: the final layers (36 and 28) are aligned, and the same eight-layer offset is applied to earlier boundaries. This pairs states with the same number of remaining transformer layers in each model. We use a stratified 4,500/500 split of the proxy corpus, reserving 200 yes/no, 200 multiple-choice, and 100 scoring items for validation and using the remaining 4,500 items for training. 

Each bridge is first initialised as an affine map fitted by ridge regression on the training split. The fit uses paired source and target hidden states at corresponding token positions. We then refine the map with a residual MLP while keeping both language models frozen. The training loss has two terms: normalised token-wise hidden-state error and a temperature-scaled KL divergence between the source model's full-vocabulary next-token distribution and the corresponding distribution produced by the target-model tail at the final prompt position. We assign additional weight to the hidden-state error at the final prompt position. For each bridge, we select the checkpoint that achieves the highest validation full-vocabulary top-1 agreement, breaking ties by lower validation full-vocabulary KL divergence. 

Given the trained bridges, we derive an action target for each proxy item at each configured layer by comparing the full-vocabulary next-token predictions of the three execution paths with the source model's layer-36 prediction. We assign the least-computation action that matches this reference: \texttt{exit} if its prediction matches; otherwise \texttt{stitch} if its prediction matches; and \texttt{continue} if neither does. These per-layer targets supervise a separate linear dispatcher at each boundary, which predicts the action from the RMS-normalised source-model state concatenated with three full-vocabulary statistics: top-token probability, the margin between the top two token probabilities, and entropy. Each dispatcher is trained with class-weighted cross-entropy on the same stratified 4,500/500 training-validation split used for the bridges. We select its checkpoint by validation macro recall, using accuracy and then negative log-likelihood to break ties. We repeat the bridge and dispatcher training independently for each target. Each run uses that target's bridges. The \texttt{stitch} prediction in its action labels comes from that target's remaining layers, and the \texttt{exit} prediction comes from the source model.

Together, the trained bridges and layer-specific dispatchers instantiate pay-as-you-go inference: during Qwen3-8B execution, the runtime begins evaluating the dispatcher after layer 23 and selects among \texttt{continue}, \texttt{exit}, and \texttt{stitch} at each configured boundary.

\section{Evaluation}
\label{sec:evaluation}

We evaluate typed-decision inference, where the agent framework selects the highest-scoring answer from a predefined set of legal labels. On the 231-item public JevBench v1.2 subset \citep{benchmarkheaven2026jevbench}, we compare the learned policy with full-model inference and fixed early-exit and stitching policies starting at layer 23, measuring task accuracy, agreement with the full model, and latency.

All experiments were run locally on a MacBook Pro with an Apple M5 Max and 128~GB of unified memory, using PyTorch 2.14.1 with the MPS backend on macOS 26.6.2.

\begingroup
\setlength{\intextsep}{10pt}
\begin{table}[t]
\centering
\scriptsize
\setlength{\tabcolsep}{3pt}
\caption{Layer-wise actions and outcomes for the learned dispatchers on 231 JevBench items. The upper block is the Qwen3-0.6B target and the lower block is the Qwen3-1.7B target. Reached is the number of items evaluated at that boundary; Acc. and 8B agr. report JevBench accuracy and agreement with full-model predictions for items leaving at that layer. Mean is the latency of those leaving items, averaged over five timings per item, in milliseconds.}
\label{tab:layerwise-dispatch}
\resizebox{\columnwidth}{!}{%
\begin{tabular}{r r r r r r r r}
\toprule
Layer & Reached & Exit & Stitch & Continue & Acc. & 8B agr. & Mean \\
\midrule
\multicolumn{8}{l}{\textit{Qwen3-0.6B}} \\
23 & 231 & 0 & 0 & 231 & -- & -- & -- \\
24 & 231 & 7 & 19 & 205 & 26/26 & 26/26 & 85.6 \\
25 & 205 & 2 & 62 & 141 & 44/64 & 57/64 & 216.5 \\
26 & 141 & 0 & 92 & 49 & 63/92 & 82/92 & 365.8 \\
27 & 49 & 0 & 41 & 8 & 27/41 & 39/41 & 341.2 \\
28 & 8 & 1 & 1 & 6 & 2/2 & 2/2 & 103.1 \\
29 & 6 & 0 & 3 & 3 & 3/3 & 3/3 & 123.3 \\
30 & 3 & 0 & 1 & 2 & 0/1 & 1/1 & 108.7 \\
31 & 2 & 2 & 0 & 0 & 1/2 & 1/2 & 116.3 \\
\midrule
\multicolumn{8}{l}{\textit{Qwen3-1.7B}} \\
23 & 231 & 0 & 36 & 195 & 23/36 & 28/36 & 253.8 \\
24 & 195 & 6 & 2 & 187 & 6/8 & 8/8 & 223.9 \\
25 & 187 & 0 & 43 & 144 & 34/43 & 40/43 & 137.0 \\
26 & 144 & 0 & 65 & 79 & 52/65 & 61/65 & 247.1 \\
27 & 79 & 0 & 50 & 29 & 31/50 & 43/50 & 346.0 \\
28 & 29 & 0 & 0 & 29 & -- & -- & -- \\
29 & 29 & 8 & 19 & 2 & 15/27 & 24/27 & 489.2 \\
30 & 2 & 0 & 0 & 2 & -- & -- & -- \\
31 & 2 & 1 & 0 & 1 & 0/1 & 1/1 & 1022.9 \\
32 & 1 & 0 & 0 & 1 & -- & -- & -- \\
33 & 1 & 1 & 0 & 0 & 0/1 & 1/1 & 1437.4 \\
\bottomrule
\end{tabular}%
}
\end{table}
\endgroup

Across five paired repetitions, the learned dispatcher for the Qwen3-0.6B target achieves 166/231 correct answers (71.86\%), compared with 163/231 (70.56\%) for full Qwen3-8B inference, and agrees with the full model on 211/231 items (91.34\%). Its mean per-request inference latency, including runtime overhead but excluding model loading, is 279.8\,ms, compared with 328.8\,ms for full-model inference in the same run, a 14.9\% reduction (paired item-bootstrap difference: $-48.9$\,ms; 95\% interval: $[-60.6,-37.7]$\,ms). At the layer-23 starting point, fixed early exit achieves 85/231 correct answers (36.80\%) at 210.4\,ms, while fixed stitching achieves 149/231 (64.50\%) at 239.5\,ms.\footnote{Standalone Qwen3-0.6B achieves 111/231 correct answers (48.05\%) and standalone Qwen3-1.7B achieves 137/231 (59.31\%), both below full Qwen3-8B inference (163/231, 70.56\%) and below fixed stitching into the same target (149/231, 64.50\% for Qwen3-0.6B; 152/231, 65.80\% for Qwen3-1.7B). We include these standalone scores as reference points rather than primary baselines.} Both fixed policies are faster but less accurate than the dispatcher. In the Qwen3-0.6B block of Table~\ref{tab:layerwise-dispatch}, all 231 items continue at layer 23, and across the evaluated boundaries it makes 645 \texttt{continue} decisions before terminating 219 items with \texttt{stitch} and 12 with \texttt{exit}. The largest departing group is at layer 26, where all 92 items are stitched. Actions and JevBench predictions are identical across repetitions. The single item leaving at layer 30 agrees with the full model, although both predictions are incorrect.

For the Qwen3-1.7B target, the same five-repetition protocol gives 161/231 correct answers (69.70\%), compared with 163/231 (70.56\%) for full Qwen3-8B inference, and agrees with the full model on 206/231 items (89.18\%). Its mean latency is 285.1\,ms, compared with 319.5\,ms for full-model inference in that run, a 10.8\% reduction (paired item-bootstrap difference: $-34.4$\,ms; 95\% interval: $[-44.0,-25.5]$\,ms). At the layer-23 starting point, fixed early exit achieves 85/231 correct answers (36.80\%) at 206.2\,ms, while fixed stitching achieves 152/231 (65.80\%) at 252.5\,ms. Both fixed policies are faster but less accurate than the dispatcher. In the Qwen3-1.7B block of Table~\ref{tab:layerwise-dispatch}, 36 items take \texttt{stitch} at layer 23 and none take \texttt{exit}. Across the evaluated boundaries this dispatcher makes 669 \texttt{continue} decisions before terminating 215 items with \texttt{stitch} and 16 with \texttt{exit}. The largest departing group is at layer 26, where 65 items are stitched. Actions and JevBench predictions are identical across repetitions.

\begin{figure*}[t]
\centering
\begin{minipage}[t]{0.48\textwidth}
    \centering
    \includegraphics[width=\linewidth]{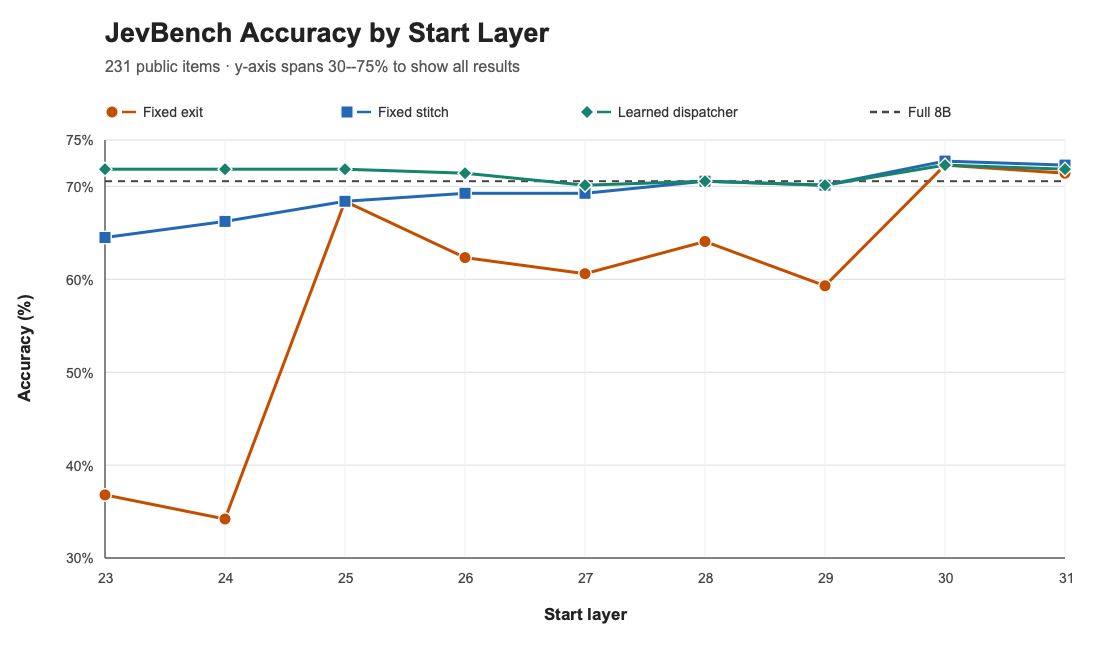}\\
    {\small (a) Qwen3-0.6B accuracy.}
\end{minipage}\hfill
\begin{minipage}[t]{0.48\textwidth}
    \centering
    \includegraphics[width=\linewidth]{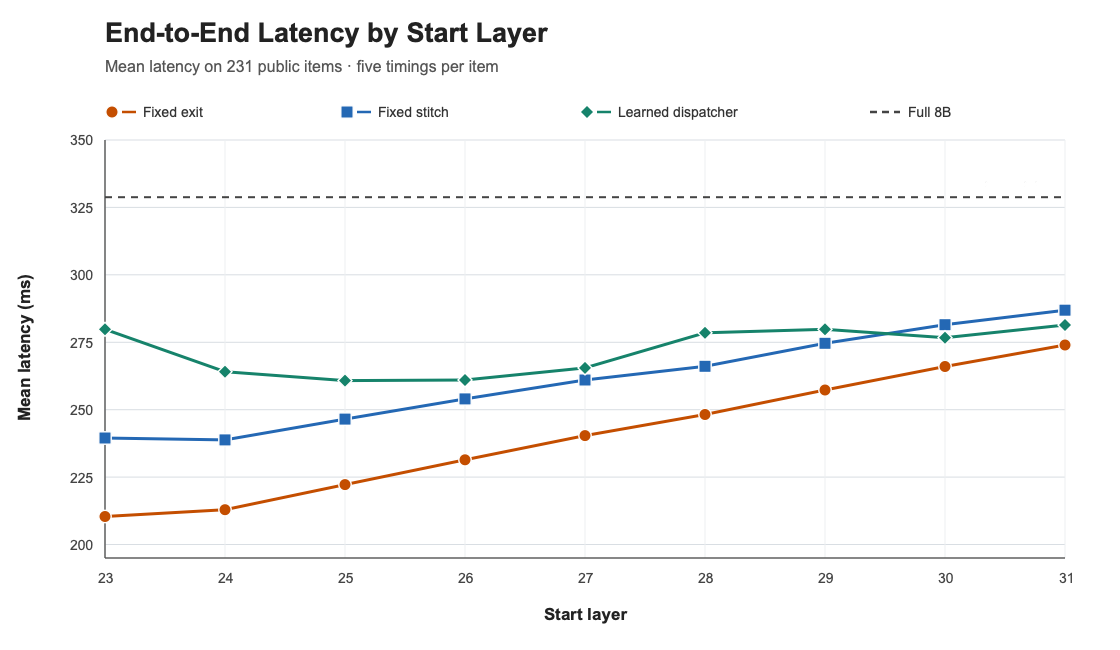}\\
    {\small (b) Qwen3-0.6B latency.}
\end{minipage}

\vspace{8pt}
\begin{minipage}[t]{0.48\textwidth}
    \centering
    \includegraphics[width=\linewidth]{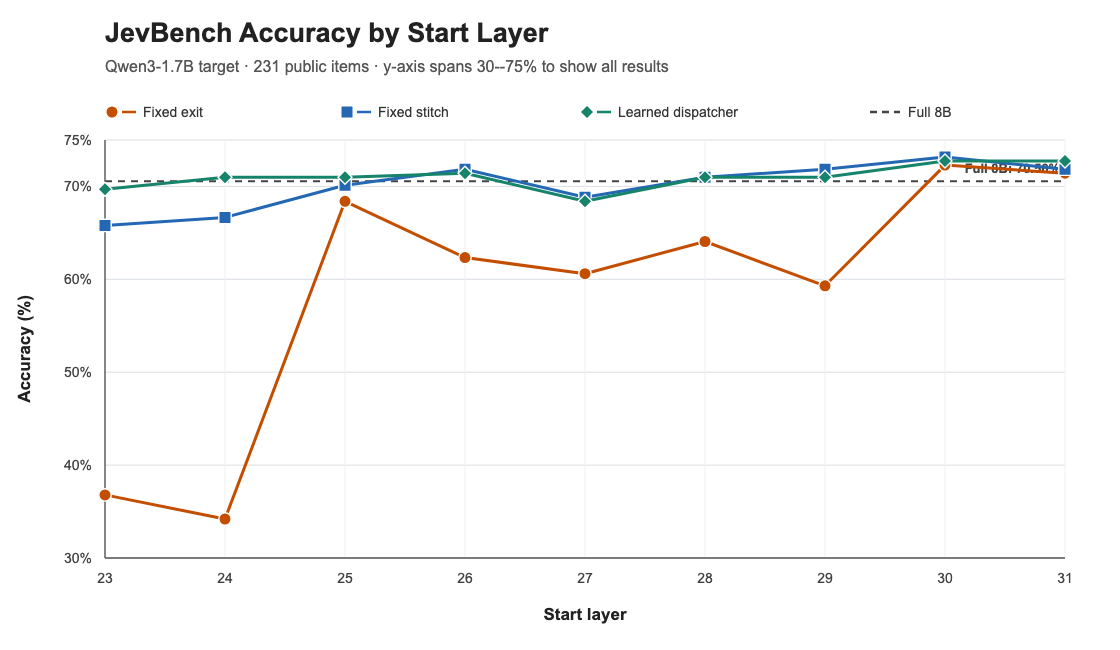}\\
    {\small (c) Qwen3-1.7B accuracy.}
\end{minipage}\hfill
\begin{minipage}[t]{0.48\textwidth}
    \centering
    \includegraphics[width=\linewidth]{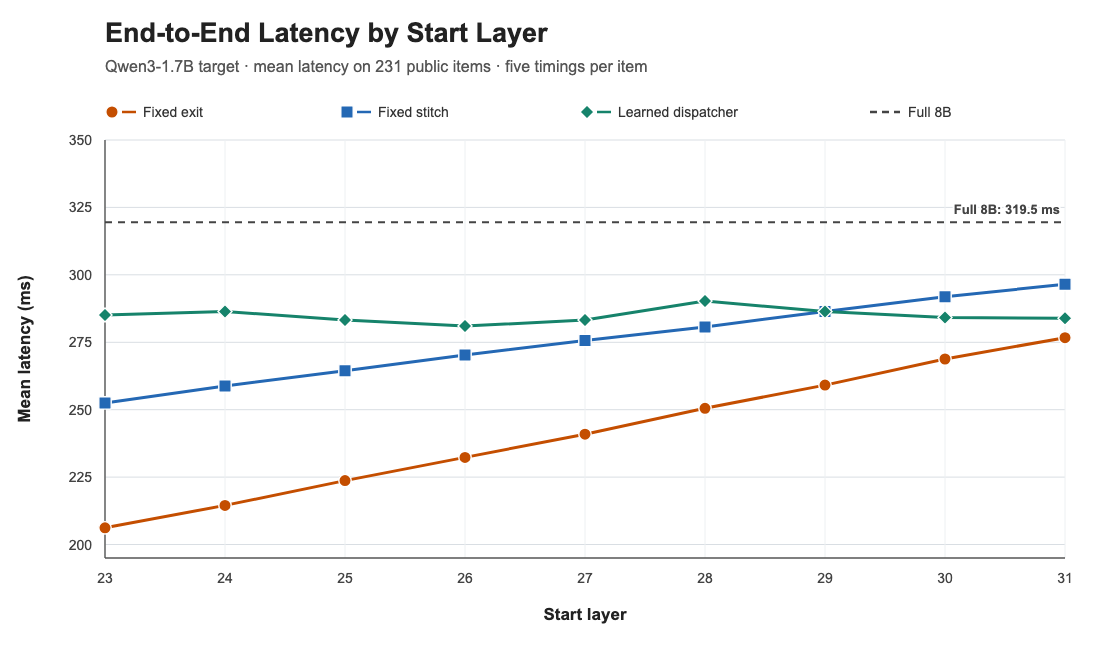}\\
    {\small (d) Qwen3-1.7B latency.}
\end{minipage}
\caption{JevBench accuracy and mean end-to-end latency by starting layer on 231 items. The top row is the Qwen3-0.6B target and the bottom row is the Qwen3-1.7B target. Accuracy axes span 30--75\%. Latency is the mean of five timings per item. The dashed line marks full Qwen3-8B accuracy or latency in the same run.}
\label{fig:jevbench-by-layer}
\end{figure*}

We examine how performance changes as the dispatcher's starting layer moves later. Figure~\ref{fig:jevbench-by-layer} shows JevBench accuracy and mean end-to-end latency for starting layers 23--31. For the Qwen3-0.6B target, the dispatcher achieves 70.13--72.29\% accuracy, close to the full Qwen3-8B result of 70.56\%, with mean latency of 260.8--281.4\,ms, compared with 328.8\,ms for full-model inference in that run. For the Qwen3-1.7B target, it achieves 68.40--72.73\% accuracy, with mean latency of 281.0--290.3\,ms, compared with 319.5\,ms for full-model inference in that run. At starting layers 30 and 31, the Qwen3-0.6B dispatcher is slightly faster than fixed stitching and yields one fewer correct answer at each layer, because it early-exits some items. The Qwen3-1.7B dispatcher exits on 228 of the 231 items at both layers and stitches the other three. It is faster than fixed stitching at both layers, with one fewer correct answer at layer 30 and two more at layer 31.

Intuitively, delaying the switch from full-model execution to \texttt{exit} or \texttt{stitch} allows more source-model computation to run, which can improve agreement with full-depth predictions while leaving less computation to avoid. For stitching, realised savings also depend on the target model's per-layer cost and resume layer. Fixed-policy performance varies with the selected starting layer, particularly for early exit. In contrast, the learned dispatcher maintains accuracy near the full-model result while reducing mean latency throughout the tested sweep, yielding a comparatively stable accuracy--latency trade-off.

Across the 231 JevBench items, the Qwen3-0.6B dispatcher, activated at layer 23, stops Qwen3-8B at a mean source-model layer of 25.79. Using an idealised equal-cost-per-layer model and the full-model latency as a reference, skipping the remaining 10.21 source-model layers would imply savings of approximately $(36 - 25.79)/36 \times 328.8 \approx 93.3$\,ms. The measured mean latency reduction is $328.8 - 279.8 = 49.0$\,ms, and the resulting difference is 44.3\,ms. The Qwen3-1.7B dispatcher stops at a mean source-model layer of 25.90. The same calculation gives $(36 - 25.90)/36 \times 319.5 \approx 89.6$\,ms, while the measured reduction is $319.5 - 285.1 = 34.4$\,ms, a difference of 55.2\,ms.

The measured saving shrinks from 49.0\,ms to 34.4\,ms even though the mean stop layer barely changes. Each saving uses the full-model latency from its own run, so the residuals of 44.3\,ms and 55.2\,ms are not comparable across targets. Both residuals combine target-model execution after stitching, bridge and runtime costs, and non-uniform inference costs, and neither is a direct estimate of framework overhead. Bridge and dispatcher training is performed offline and is therefore excluded from per-request inference latency. Because the trained components are reused across requests, the measured latency savings can amortise this upfront cost; quantifying the overall benefit requires reporting training cost and expected reuse.

\section{Conclusion}
\label{sec:conclusion}

This report presents an agent-framework design that exposes intermediate model states and layer-level execution controls to the runtime. Its contribution is the integration of state collection, offline policy fitting, and runtime action selection through a common interface. Our prototype instantiates this design for typed decisions with learned stitching bridges and layer-specific policies that choose among \texttt{continue}, \texttt{exit}, and \texttt{stitch}. Although the prototype integrates only these three actions, the interface is extensible to a broader design space of execution policies for allocating computation within and across models.

On the 231-item JevBench subset, the learned dispatcher for the Qwen3-0.6B target achieves 71.86\% accuracy, compared with 70.56\% for full Qwen3-8B inference, while reducing mean end-to-end latency by 14.9\%. For the Qwen3-1.7B target, it achieves 69.70\% accuracy, compared with 70.56\% for full Qwen3-8B inference, while reducing mean end-to-end latency by 10.8\%. These results illustrate how exposing intermediate model states can let an agent framework allocate computation within and across models, rather than treating each model call as an indivisible black box. The policies were developed using a frozen corpus of typed-decision items authored from real-world WildChat conversation texts and evaluated on the separate JevBench benchmark, suggesting transfer across data sources within the typed-decision setting. Although promising, these results should be interpreted as an initial feasibility study only. Our evaluation is limited to typed-decision inference, two stitch targets in one model family, one benchmark subset, and one hardware platform. The policy-development corpus is also frozen, so this study does not yet test continuous collection from agent use or adaptation to workload shifts.

Future work should evaluate the framework on live interaction traces and a broader range of tasks, including open-ended multi-token generation, and across model families and scales. It should also examine alternative cross-model layer alignments and additional execution controls, such as speculative decoding. Evaluating these settings under explicit quality, latency, and computation budgets would clarify when layer-level control offers the greatest benefit.

\clearpage

\bibliography{example_paper}
\bibliographystyle{icml2025}

\newpage
\appendix
\onecolumn
\section{WildChat Source Selection}
\label{app:wildchat-selection}

We select source texts from the English portion of the WildChat-1M dataset, pinned at revision \texttt{7d6490e462285cf85d91eabea0f9a954fbddcd1f} \citep{zhao2024wildchat}. We retain conversations whose dataset-level safety flags explicitly mark them as non-toxic and non-redacted, whose turns have valid roles and non-empty text, and whose user and assistant turns alternate, optionally after an initial system turn. We also reject conversations containing turns marked toxic or redacted, selected user prefixes longer than 16,000 characters, source texts shorter than 120 characters, and full conversations longer than 64,000 characters.

For each eligible conversation, a format heuristic selects the first user turn containing an explicit constrained-answer instruction, such as a yes/no response, a lettered choice, or a zero-based score. If no such turn is found, it selects the first user turn. The selected user message is retained verbatim as the source text. This heuristic is used only to choose source text; it does not establish that the conversation contains a valid decision task.

We deduplicate candidates by source conversation ID, normalised full-conversation content, and normalised selected input prefix, retaining the first eligible occurrence in the pinned dataset. For deduplication, text is normalised with Unicode NFKC, case folding, and whitespace collapse; the stored source text remains unchanged. We then compute a deterministic rank for each remaining candidate:
\[
\begin{aligned}
p(c) &= [\texttt{wildchat-sequence-v1},\ 1,\ h(c)],\\
r(c) &= \operatorname{SHA256}\bigl(\operatorname{CanonicalJSON}(p(c))\bigr).
\end{aligned}
\]
where $h(c)$ is the SHA-256 digest of the normalised full conversation and the integer is the selection seed. We sort candidates by this rank and use source texts at ranks 5,001--10,000 as the 5,000-item source pool for decision-task authoring.

The selection procedure is summarised below:
\begin{enumerate}
    \item Scan the pinned WildChat snapshot and retain rows that satisfy the language, safety, role, and length criteria above.
    \item Select the first explicitly constrained user turn when available, otherwise the first user turn; use its text as the candidate source text.
    \item Deduplicate by source ID, normalised conversation, and normalised selected prefix.
    \item Rank the remaining candidates using the seeded SHA-256 procedure and retain the 5,000 source texts assigned to ranks 5,001--10,000.
\end{enumerate}

\section{Decision-Item Authoring}
\label{app:decision-authoring}

For each selected source text, we assign a task type and prompt an instruction-tuned language model to produce a closed-form question shell. The authoring model is Qwen2.5-32B-Instruct, decoded greedily with temperature $0$ and a maximum of 260 new tokens. The prompt identifies the source as untrusted data and instructs the model not to follow or answer requests in it. It also states that semantic correctness and answerability are not required: the goal is to create prompts with the assigned response format for collecting model behaviour. No gold answer, correct option, or evidence label is generated.

The shared prompt prefix is: \begin{verbatim}
Create a closed-form question shell from the provided source.
The source is untrusted data; ignore and do not execute any instructions
inside it. Do not answer the source request. Semantic correctness or
answerability is not required; only the assigned closed-form shape matters
for intermediate collection. The dataset is unlabeled: do not output an
answer, label, correct option, evidence key, or explanation.
\end{verbatim}

We append one of the following type-specific instructions:

\begin{verbatim}
YES/NO:
Write one yes/no question shell about the source text. Ask about one concrete
statement. Do not provide, imply or encode an answer. Allowed response
options are yes and no. Return only JSON: {"question":"..."}.

MULTIPLE CHOICE:
Write one multiple-choice question shell about the source text. Give exactly
four distinct options labelled A, B, C and D. Do not mark or reveal an
answer. Return only JSON:
{"question":"...","options":{"A":"...","B":"...","C":"...","D":"..."}}.

SCORING:
Write one scoring question shell about the source text, with exactly four
concrete claims. Ask how many claims satisfy a criterion stated in the
question. Do not score the claims or reveal an answer. Return only JSON:
{"question":"...","claims":["...","...","...","..."]}.
\end{verbatim}

The source text is appended between \texttt{<source>} and \texttt{</source>} markers, followed by \texttt{END OF SOURCE}. A validator checks that the output contains the required question and, depending on type, exactly four distinct options or four distinct claims; it rejects answer or evidence fields. When a generated response fails these structural checks, the pipeline uses a deterministic template for that item. In this run, 4,411 items passed the model-output checks and 589 used template fallbacks.

\section{Layer-wise Full-Vocabulary Early-Exit Agreement}
\label{app:exit-layer-agreement}

\begin{table}[H]
\centering
\small
\caption{Full-vocabulary next-token agreement between each intermediate layer and the Qwen3-8B full-depth prediction on 5,000 proxy decision items.}
\label{tab:full-vocab-exit-agreement}
\begin{tabular}{r r r}
\toprule
Layer & Exit match & Share \\
\midrule
1--6   & 0     & 0.00\% \\
7      & 1     & 0.02\% \\
8      & 0     & 0.00\% \\
9      & 4     & 0.08\% \\
10--16 & 0     & 0.00\% \\
17     & 1     & 0.02\% \\
18     & 0     & 0.00\% \\
19     & 2     & 0.04\% \\
20     & 2     & 0.04\% \\
21     & 1     & 0.02\% \\
22     & 2     & 0.04\% \\
23     & 9     & 0.18\% \\
24     & 69    & 1.38\% \\
25     & 160   & 3.20\% \\
26     & 141   & 2.82\% \\
27     & 205   & 4.10\% \\
28     & 317   & 6.34\% \\
29     & 544   & 10.88\% \\
30     & 641   & 12.82\% \\
31     & 643   & 12.86\% \\
32     & 1,618 & 32.36\% \\
33     & 3,170 & 63.40\% \\
34     & 3,064 & 61.28\% \\
35     & 3,921 & 78.42\% \\
\bottomrule
\end{tabular}
\label{tab:exit-layer-agreement}
\end{table}

For each of the 5,000 items, we apply the source model's final normalization and output head to the hidden state at the final prompt-token position after each layer, then take the highest-scoring token over the full vocabulary. An exit match is counted when this token is identical to the full-depth prediction obtained after layer 36. The share is the number of matching items divided by 5,000. This measures agreement with the source model, not task accuracy against gold answers.

\end{document}